\documentclass[letterpaper, 10 pt, conference]{ieeeconf} 
\IEEEoverridecommandlockouts                              
\usepackage{times}
\usepackage{multicol}
\usepackage[bookmarks=true]{hyperref}
\usepackage[table, dvipsnames]{xcolor}
\usepackage{amsmath}
\usepackage{amssymb}
\usepackage{bm}
\usepackage{graphicx}
\usepackage{dsfont} 
\usepackage{float}
\usepackage{subfigure}
\usepackage{cleveref}
\usepackage[normalem]{ulem}
\usepackage{adjustbox}
\usepackage{booktabs}
\usepackage{algorithm}
\usepackage{algpseudocode}
\usepackage{cite}

\newcommand{\mbb}[1]{\mathbb{#1}}

\newcommand{\T}{\mathsf{T}}

\title{\LARGE \bf
\textbf{VISTA}: Open-Vocabulary, Task-Relevant Robot Exploration with Online Semantic Gaussian Splatting
}

\author{Keiko Nagami$^{1}$, Timothy Chen$^{1}$, Javier Yu$^{1}$, Ola Shorinwa$^{1}$, Maximilian Adang$^{1}$, \\ Carlyn Dougherty$^{2}$, Eric Cristofalo$^{2}$, and Mac Schwager$^{1}$
\thanks{$^{1}$Department of Aeronautics and Astronautics,
        Stanford University, Stanford, CA 94305, USA
        {\tt\small knagami, chengine, javieryu, shorinwa, madang, schwager@stanford.edu}.}%
\thanks{$^{2}$MIT Lincoln Laboratory, Lexington, MA 02421, USA
        {\tt\small carlyn.dougherty, eric.cristofalo@ll.mit.edu}.}%
\thanks{This work was supported in part by MIT Lincoln Laboratory ACC grant 7000603941, NSF project FRR 2342246, and ONR project N00014-23-1-2354. The NASA University Leadership initiative (grant \#80NSSC20M0163) provided funds to assist the authors with their research, but this article solely reflects the opinions and conclusions of its authors and not any NASA entity.}%
\thanks{\tiny DISTRIBUTION STATEMENT A. Approved for public release. Distribution is unlimited.
This material is based upon work supported by the Department of the Air Force under Air Force Contract No. FA8702-15-D-0001 or FA8702-25-D-B002. Any opinions, findings, conclusions or recommendations expressed in this material are those of the author(s) and do not necessarily reflect the views of the Department of the Air Force.
\textcopyright 2025 Massachusetts Institute of Technology.
Delivered to the U.S. Government with Unlimited Rights, as defined in DFARS Part 252.227-7013 or 7014 (Feb 2014). Notwithstanding any copyright notice, U.S. Government rights in this work are defined by DFARS 252.227-7013 or DFARS 252.227-7014 as detailed above. Use of this work other than as specifically authorized by the U.S. Government may violate any copyrights that exist in this work.}%
}

\begin{document}

\maketitle
\thispagestyle{empty}
\pagestyle{empty}

\begin{abstract}
    We present VISTA (Viewpoint-based Image selection with Semantic Task Awareness), an active exploration method for robots to plan informative trajectories that improve 3D map quality in areas most relevant for task completion. Given an open-vocabulary search instruction (e.g., ``find a person"), VISTA enables a robot to explore its environment to search for the object of interest, while simultaneously building a real-time semantic 3D Gaussian Splatting reconstruction of the scene. The robot navigates its environment by planning receding-horizon trajectories that prioritize semantic similarity to the query and exploration of unseen regions of the environment. To evaluate trajectories, VISTA introduces a novel, efficient viewpoint-semantic coverage metric that quantifies both the geometric view diversity and task relevance in the 3D scene. 
On static datasets, our coverage metric outperforms state-of-the-art baselines, FisherRF and Bayes' Rays, in computation speed and reconstruction quality. In quadrotor hardware experiments, VISTA achieves 6x higher success rates in challenging maps, compared to baseline methods, while matching baseline performance in less challenging maps. Lastly, we show that VISTA is platform-agnostic by deploying it on a quadrotor drone and a Spot quadruped robot. Open-source code will be released upon acceptance of the paper.
\end{abstract}

\section{Introduction}
\label{sec:introduction}

\begin{figure}[t]
    \centering
    \includegraphics[width=\linewidth, trim={0 2cm 0 0}]{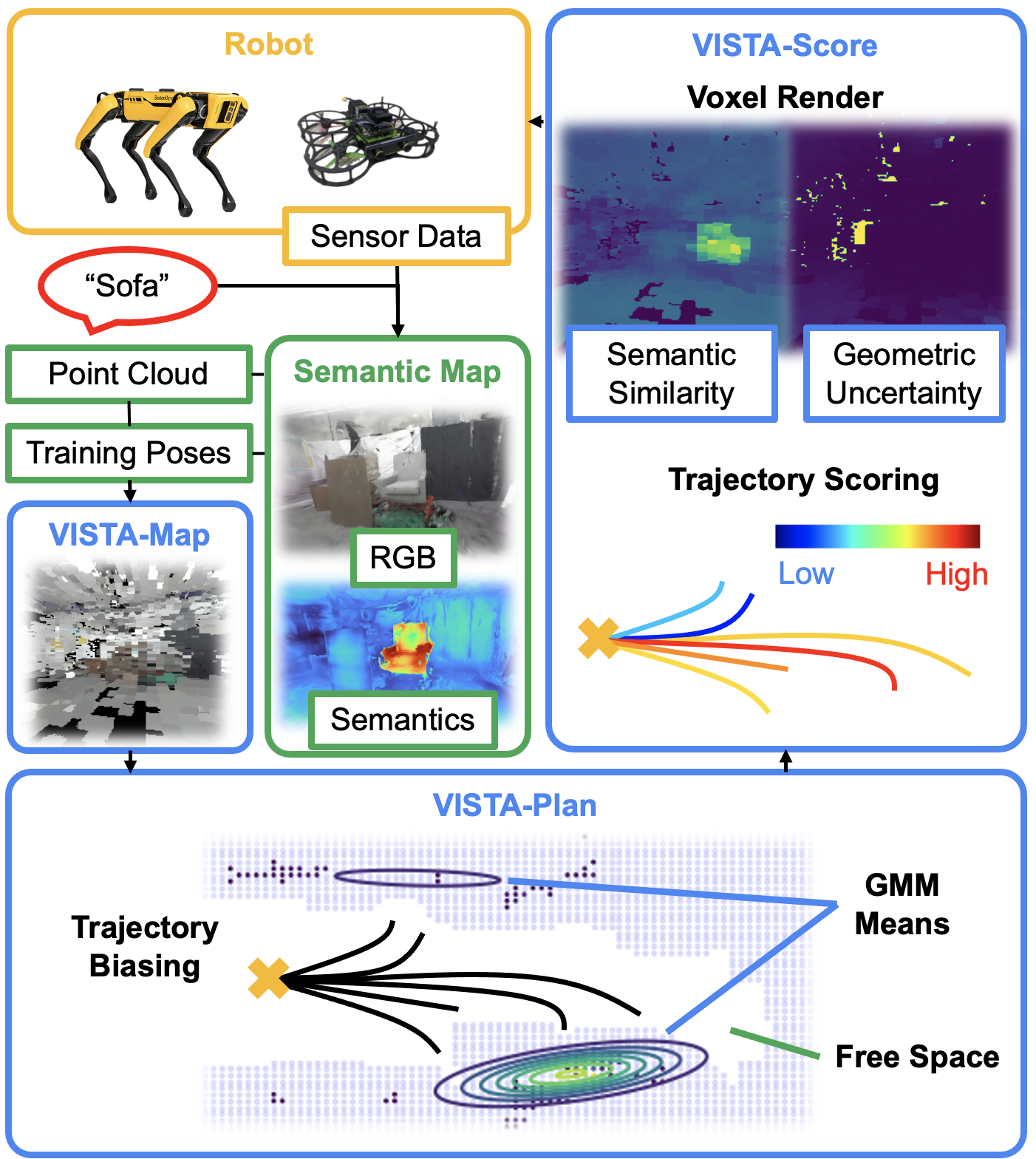}
    \caption{System overview of VISTA. Real-time sensor data is gathered from a robot hardware platform to train a semantic 3DGS map. The semantic and RGB information from the 3DGS map are transferred to a 3D Voxel Grid, and training poses are used to store geometric information about directions from which each voxel has already been viewed. In the planner, the 3D voxel grid is flattened into a top-down 2D voxel grid, where frontier cells and semantic information are used to fit a Gaussian Mixture Model that is sampled to generate candidate trajectories. The trajectory with the highest semantic + geometric information gain is then executed in a receding horizon loop.}
    \vspace{-1em}
    \label{fig:system}
\end{figure}

Research advances in vision and language foundation models, e.g., \cite{radford2021learning, caron2021emerging}, have enabled language-guided object localization in pre-mapped real-world environments \cite{firoozi2023foundation}. For example, a user can task a robot with the word \emph{apple} and it will find an apple in a pre-mapped concept graph \cite{gu2024conceptgraphs}. However, to find query objects efficiently in \emph{unstructured, unmapped environments}, robots must be capable of exploring their environments intelligently, with a bias toward finding the object of interest. Prior work in robot exploration broadly uses traditional $3$D scene representations, such as occupancy grids and voxel grids. We build upon these traditional representations by introducing a Gaussian Splat embedded with semantic information that can advance downstream tasks to be performed by the robot. 

We present VISTA, an algorithm for \textbf{V}iewpoint-based \textbf{I}mage Selection with \textbf{S}emantic \textbf{T}ask \textbf{A}wareness. To enable task-relevant exploration, VISTA introduces two key innovations: (i) a semantics-aware mapping and information-gain pipeline that leverages open-vocabulary semantics for task-relevant exploration, and (ii) a scalable information-gain metric based on view angle diversity that can be computed efficiently in real-time.  Our key insight is that, for vision-based mapping, the variety and multiplicity of viewing directions from which an environment point has appeared in the image history is a strong proxy for the geometric quality of the reconstruction at that point.

First, VISTA builds a high-fidelity photorealistic map of the robot's environment online using a Gaussian Splatting (3DGS) representation \cite{kerbl20233d}.\footnote{VISTA can equivalently use a neural radiance field (NeRF) \cite{mildenhall2021nerf} or any other high fidelity scene representation that can embed semantic codes.} To enable open-vocabulary, task-relevant robot exploration, VISTA distills semantic features from vision-language models, e.g., CLIP \cite{radford2021learning}, into the 3DGS map incrementally as new observations are obtained by the robot. The semantic features encode the relevancy between each point in the environment and the specified exploration task, giving a task relevancy 3D heatmap. 

Using the 3DGS map, VISTA simultaneously constructs a voxel grid, capturing the viewed regions of the scene and the semantic relevance of these regions. With this grid, we measure view diversity through a conceptually simple, computationally efficient coverage metric with advantages like recursive updating and informative trajectory optimization.  In each voxel, the geometric uncertainty is the minimum angular separation between the test viewpoint and all view angles from which that voxel has appeared in the image history accounting for occlusions. 

Finally, VISTA samples trajectories, and selects those with viewpoints that maximize a weighted combination of geometric uncertainty and semantic relevance, ultimately guiding the robot toward a specified query object. We illustrate these components in \Cref{fig:system}. Through an experimental campaign with a total of $36$ hardware executions, we show that VISTA outperforms state-of-the-art baselines, achieving 6x better success rates in environments where the object is not visible from the initial robot pose, and matching performance when it is visible.

Our contributions are as follows. We introduce:
\begin{enumerate}
    \item an efficient information metric that combines view angle diversity and semantic task relevance stored on a voxel grid that can be recursively updated, 
    \item a real-time informative trajectory planning algorithm to drive robot exploration using this metric,
    \item and a full stack ROS implementation of VISTA and online 3DGS training demonstrated on robot hardware.
\end{enumerate}
\section{Related Work}
\label{sec:related_work}
\smallskip

\noindent\textbf{Robot Exploration.}
The objective in robot exploration is to traverse through an environment efficiently to build a map, without colliding into obstacles. The maps used in these problems have traditionally used occupancy grids, point clouds, signed distance fields (SDFs), and voxel maps \cite{elfes1989using, kim2018slam, ryde20103d}. Within these map environments, exploration is typically performed through frontier-based, and viewpoint sampling methods. Frontiers are defined within the map as boundaries between free and unknown space. The robot then plans to the frontiers to collect observations of the unknown regions \cite{yamauchi1997frontier, yu2023echo, cieslewski2017rapid}. In viewpoint sampling methods, each view is scored by an information metric where the highest scoring are prioritized destinations \cite{bircher2018receding, saulnier2020information}. These two methods are often used together, in order to improve sample efficiency of the candidate viewpoints \cite{tao2023seer, kompis2021informed, jiang2023fisherrf}.

Building upon these methods, VISTA incorporates both frontier-based and viewpoint sampling methods for its exploration pipeline, using a high-fidelity 3DGS map representation, while simultaneously incorporating semantic information in both its map and trajectory scoring to search for objects in the scene.

\noindent\textbf{Active Planning and View Selection in Radiance Fields.}
Recently, radiance field methods, such as NeRFs \cite{mildenhall2021nerf, mipnerf360, mueller2022instant} and 3DGS  \cite{kerbl20233d, yu2024mip} have been introduced to robot mapping \cite{nerf-slam, yan2024gs, matsuki2024gaussian}, to generate high fidelity representations of environments. Of these methods, 3DGS provides much faster training and rendering rates, enabled by an explicit, interpretable representation of the scene, and a tile-based rasterization procedure, which is more efficient than the volumetric ray-marching procedure used by NeRFs.

Active planning algorithms use view selection methods in order to gather the most information. Several methods for radiance fields \cite{zhan2022activermap, lee2022uncertainty, yan2023active_neural, yan2023active_implicit, jin2024gs, xu2024hgs, jiang2024ag, jin2025activegs} perform active view planning to improve localization accuracy while reducing the risk of collision in navigation. A notable body of work \cite{liu2024beyond, xu2024hgs, jiang2024ag} leverage the Fisher information of radiance fields to estimate the uncertainty of 3DGS models, optimizing over a trajectory of viewpoints to maximize the information gain while minimizing localization errors \cite{jiang2024ag, liu2024beyond}. In \cite{goli2024bayes}, the authors use a similar mechanism to quantify epistemic uncertainty in NeRFs, however these methods are not used directly for active mapping. In RT-GuIDE \cite{tao2024rt}, the authors compute the magnitude of the change in the parameters of a 3DGS map over consecutive updates to estimate the uncertainty of the map in active exploration, and showcase demonstrations of their approach on a robot hardware platform.

Many of these methods rely on the radiance field map to obtain information metrics, and as such are not easily transferrable to other map representations. Additionally, none of these methods account for semantic information in the map, nor do they reason about task-relevance in planning.

\noindent\textbf{Semantics-Guided Exploration.}
The methods used for active planning and view selection focus largely on improving the quality of the map, and typically do not reason about any specified task. In this work, we not only hope to obtain a well constructed map, but additionally require our robot to find specific objects queried by a user. 2D vision language foundation models enable distillation of information into 3D scene geometry. Specifically, semantic radiance fields \cite{kerr2023lerf, zhou2024feature, qin2024langsplat, shorinwa2024fast, yu2025hammer, shorinwa2025siren}, capture $2$D semantic information from vision foundation models, e.g., CLIP \cite{radford2021learning}, DINO \cite{caron2021emerging}, SAM \cite{kirillov2023segment}, and LLaVA \cite{liu2024visual} into maps expressing semantics in $3$D space. 

Several works address the use of semantic maps for object search, \cite{gadre2023cows, yokoyama2024vlfm, ong2025atlasnavigatoractivetaskdriven, papatheodorou2023finding, barbas2025findanything}. In \cite{gadre2023cows}, the authors compare a number of different exploration methods, combined with object detection modules that are used to define a switch from exploration to exploitation. However, none of these methods are performed on robot hardware platforms. In \cite{yokoyama2024vlfm}, the authors also use a switching mode from exploration to exploitation, and present a novel geometry-based metric to quantify the confidence in the map to represent their information gain to select frontiers for exploration. In \cite{ong2025atlasnavigatoractivetaskdriven}, the authors present a method for active search with task specification on large-scale maps by using smaller local 3DGS submaps as the robot moves through a global scene. The paper demonstrates the method used on real robot platforms in multiple different maps in both indoor and outdoor environments. While the results show great versatility in multiple environments, the method does not account for geometric uncertainty in the map, so it is unclear if this method would be able to perform well if the query object were not in the initial map.
In contrast to these methods, VISTA integrates both semantics and geometric guidance into the exploration task robustly without relying on switching behavior modes, addressing these limitations.

\section{Problem Formulation}
\label{sec:problem}
We consider a robotic exploration problem in which a robot has an onboard, forward-facing RGB-D camera with reliable state estimation. The robot is placed into a previously unseen environment and is given an open-vocabulary query to locate and retrieve a certain object in the scene by a user. Once the robot receives the input query, it must then construct a map of its environment as it moves, while simultaneously searching for the query object. In this informative planning task, the robot must balance the requirement of finding the object while generating a map during exploration to confidently determine whether or not the identified object satisfies the user's input query.

To train the 3DGS and render images in the voxel grid, the camera pose of the robot's onboard camera is represented as: 
${\textbf{x} = \begin{bmatrix}
       x & y & z & \phi & \theta & \psi
   \end{bmatrix}^T,}$
representing the position and Euler angles of the camera in the global frame. As the robot moves, it collects full pose odometry information along with RGB and depth images in order to train a 3DGS map of the environment. 3DGS \cite{kerbl20233d} represents non-empty space using Gaussian primitives, each parametrized by a mean (center) ${\boldsymbol{\mu} \in \mbb{R}^{3}}$, a covariance ${\boldsymbol{\Sigma} = \textbf{R}\textbf{S}\textbf{S}^{\T}\textbf{R}^{\T} \in \mathcal{S}^n_{++}}$ (parameterized by a rotation matrix $\textbf{R}$ and a diagonal scaling matrix $\textbf{S}$), an opacity ${\alpha \in \mbb{R}_{+}}$, and spherical harmonics (SH) parameters to capture view-dependent visual effects like reflections. This explicit representation not only enables Gaussian Splatting to avoid unnecessary computation involving empty space, but it also enables the utilization of fast tile-based rasterization. The rasterization procedure uses $\alpha$-blending, computing the color of each pixel.

As the map updates, we assume that the motion of the robot is restricted in the $z$, $\phi$, and $\theta$ axes. The robot's motion is then modeled as a planar single integrator with a heading angle in the yaw direction. The state and control vectors for planning ${\textbf{s} \in \mbb{R}^{3}}$ and ${\textbf{u} \in \mbb{R}^{3}}$ are as follows:
\begin{align}
   \textbf{s} &= \begin{bmatrix}
       x & y & \psi
   \end{bmatrix}^T, \quad
   \textbf{u} = \begin{bmatrix}
       \dot{x} & \dot{y} & \dot{\psi}
   \end{bmatrix}^T,
\end{align}
where the velocity control vector is subject to control limits. 

\section{VISTA}
\label{sec:method}
We propose a method for efficient exploration of an environment by a robot, guided by a natural language semantic query provided by a user. Our robot constructs a 3DGS map in real-time as it moves, collecting odometry and RGB-D image information. We extract a voxel grid from the underlying radiance field for computational efficiency when computing information gain metrics. Within this voxelized representation, we distinguish between observed and unobserved space, where the observed space is further segmented into occupied and free space. To enable task-aligned scene coverage, we fuse geometric and semantic information-gain metrics, guiding the robot toward regions with high geometric uncertainty and semantic relevancy to the input query.
Our method is visualized in Fig. \ref{fig:system}.

\subsection{Real-Time Semantic Radiance Field Training}
To construct semantic 3DGS maps in real-time, VISTA builds upon NerfBridge \cite{yu2023nerfbridge, nerfstudio} and its 3DGS extension Splatbridge \cite{chen2024safer}, both real-time methods for online training of radiance fields. In both,  images from the robot's onboard cameras and poses are aggregated into a streaming dataset that is used to continuously optimize the radiance field. To add semantic information to these platforms, we then distill semantic embeddings from the $2$D vision-language model CLIP \cite{radford2021learning}
into the online radiance field using the distillation procedure used in semantic NeRF literature \cite{shen2023F3RM}, computing the CLIP image embeddings for each incoming image. Specifically, the trained semantic field ${f: \mbb{R}^3 \mapsto \mbb{R}^{l}}$, parametrized by a multi-resolution hashgrid followed by a multilayer perceptron, maps a $3$D point to a semantic embedding.

To optimize the semantic field, $3$D points back-projected from the predicted depth image are used to generate inputs for training $f$. The parameters of $f$ are optimized using gradient-based optimization of the mean-squared error (MSE) between the predicted and ground-truth CLIP semantic embeddings. We also include the cosine similarity as a component in the loss function in 3DGS. The parameters of $f$ and those of the base 3DGS are trained simultaneously.

While the 3DGS map is training, a point cloud representation of the field containing RGB colors and semantic embeddings is published in real-time. Simultaneously, a subset of the training poses is also returned. If the dataset size is below some maximum number of poses $N$, the full dataset is returned, otherwise $N$ poses are sampled from the training dataset. 

\begin{figure}[th]
    \centering
    \includegraphics[width=\columnwidth, clip, trim={1.5em 3em 1.5em 1.5em}]{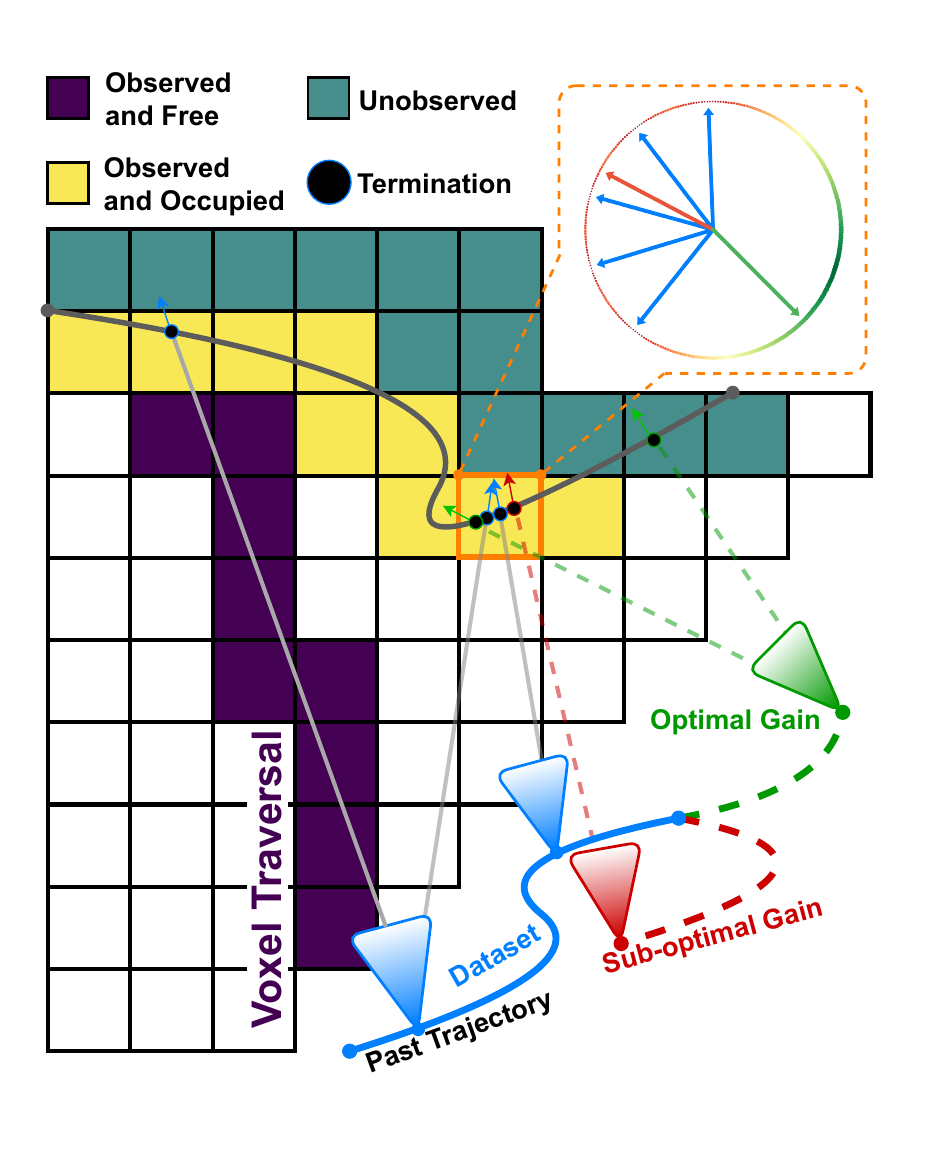}
    \vspace{-1.5em}
    \caption{Geometric information gain based on view diversity coverage. Given a point cloud, voxels are characterized as free, occupied, and unobserved. These categories are determined through voxel traversal over the camera rays. All occupied voxels contain a list of view directions derived from the dataset (blue). Pixel-level geometric gain can be rendered from arbitrary viewpoints  through voxel traversal, terminating at either occupied or unobserved voxels. A simple coverage metric is computed between the candidate view direction and the per-voxel list of dataset-derived directions. Rays that hit unobserved voxels always render the highest gain. The gain metric first prioritizes viewing unobserved regions and then viewing occupied regions from different directions. As a result, viewpoints that are similar to the dataset render to a sub-optimal gain (red), while those that are  different return a high gain (green). The mechanism is visualized in the inset, with low gain directions being red and pointing to denser areas of the unit sphere, while high gain directions are green and point towards sparser regions.}
        \vspace{-1em}
    \label{fig:occupancy_grid}
\end{figure}

\subsection{VISTA-Map: 3D Voxel Grid Representation}
\label{sec:vista-map}
The published point cloud from the 3DGS training procedure is voxelized into a grid of fixed size and resolution centered on the robot's current position. In this way, the map representation is restricted from growing with the number of Gaussians in the 3DGS map. 

The voxels in the grid are characterized into one of three categories: occupied, free space, and unobserved. All point cloud information (e.g. color and semantic embeddings) is registered into occupied voxels. In order to separate the remaining voxel grid cells into either unobserved or free cells, VISTA employs voxel traversal \cite{amanatides1987voxeltraversal} from the training cameras of the 3DGS pipeline. All rays corresponding to pixels of the training camera originate at the camera origin and either terminate at an occupied voxel or exit the voxel grid. All voxels that intersect with rays up until termination are deemed free space. The remaining voxels are assigned as unobserved voxels. 

Voxel traversal of the voxel map can similarly be used to render RGB, depth, semantic, and geometric gain images. From camera pose $\textbf{x}$, rays parametrized as $\textbf{r}(t) = \textbf{o} + t\cdot \textbf{d}$ corresponding to each pixel are cast into the voxel map, terminating when either an occupied or unobserved voxel is traversed or some maximum draw distance is reached. Rays that terminate at an occupied or unobserved voxel render that voxel's attributes into the pixel. The geometric gain and semantic images are used in active view planning, detailed in Section \ref{sec:vista-score}.

\subsection{VISTA-Score: Information Gain Quantification}
\label{sec:vista-score}

Once the voxel grid representation of the environment is obtained, we compute information gain metrics from the voxel grid. As described in Section \ref{sec:vista-map}, the training poses and the voxel traversal mechanism are used to store information about the directions from which camera rays of the training views intersect with the respective voxel.

With this stored information, we evaluate the proposed geometric information gain of a candidate camera pose for a particular voxel by comparing the new proposed view direction with the existing view directions for that voxel. Specifically, we compute the dot product of each of the existing voxel's direction vectors against the new camera pose rays that intersect with that voxel. For new rays that contrast greatly from the closest training ray, the dot product will approach $-1$. For new rays that are very similar to the closest training ray, the dot product will approach $1$. We then normalize this information between $0$ and $1$ for each ray from the proposed camera for the following pixel-wise gain metric,

\begin{align}
    g_{\mathcal{I}}(\textbf{d}_{\textbf{x}}^n) &= \frac{\min(-\textbf{d}_{v}^T \textbf{d}_{\textbf{x}}^n) + 1}{2},
\end{align}
where $\textbf{d}_{\textbf{x}}$ is an array of all direction vectors of rays generated from camera pose $\textbf{x}$, and the superscript $n$ denotes a single ray index in this array.  Similarly, $\textbf{d}_{v}$ is an array of all direction vectors $v$ stored in that voxel from previous camera views. The $\min$ is taken over indices of the resulting vector, to give the cosine similarity between the proposed view and closest existing view.  Subsequently, we compute an image-wise geometric information gain for the candidate pose by taking the mean of the pixel values,
\begin{align}
    G_{\mathcal{I}}(\textbf{x}) &= \frac{1}{N_r}\sum_{n = 1}^{N_r} g_{\mathcal{I}}(\textbf{d}_\textbf{x}^n),
\end{align}
where $N_r$ is the number of rays from the image taken at pose $\textbf{x}$. For the semantic information gain, we also use the voxel traversal mechanism to produce a per-pixel semantic value in images rendered in the voxelized map.  The semantic information gain is then computed as the mean value of all pixels in the image, denoted by $G_{\mathcal{S}}(\textbf{x})$, and the two metrics can be used together to compute a VISTA-Score along a trajectory of viewing directions as follows,
\begin{align}
    \label{eq:vista-score}
    G(\bar{\textbf{x}}) &= \sum_{\textbf{x} \in \bar{\textbf{x}}} \gamma^{K-k} (c G_{\mathcal{I}}(\textbf{x}) + G_{\mathcal{S}}(\textbf{x})),
\end{align}
where the path $\bar{\textbf{x}}$ is described by a sequence of waypoints $\textbf{x}$, $K$ is the number of waypoints in the path, $k$ is the index of the waypoint in the path, $\gamma$ is a discount factor, and $c$ is a weighting factor of the geometric information gain. In Section \ref{sec:vista-plan}, we detail how paths are sampled for scoring.

\subsection{VISTA-Plan: Informative Planning}
\label{sec:vista-plan}
To generate candidate paths for exploration (\ref{eq:vista-score}), we sample trajectories that are biased toward regions of high information gain. The algorithm used for planning is detailed in Algo. \ref{alg:vista-plan}. In this planning pipeline, we first create a 2D voxel grid, $\mathcal{V}'$ from the 3D voxel grid, $\mathcal{V}$, by slicing a band of $\mathcal{V}$ in the $z$-direction in which the robot will operate, omitting the floor and ceiling. The voxels are then assigned in $\mathcal{V}'$ by priority in order of observed-occupied, unobserved, then observed-free, along the $z$-dimension. Semantic information is similarly encoded in $\mathcal{V}'$, by slicing a band of the 3D voxel grid, and summing the semantic values of each voxel grid cell in the height dimension. In Algo. \ref{alg:vista-plan}, this procedure is captured in the function \texttt{FlattenVoxelGrid}.

Using this 2D representation of the environment, we encode global geometric and semantic information from the scene to bias trajectories toward regions of the environment with highest information gain. For geometric information gain, we search for frontier cells, $\mathcal{D}_f$ from $\mathcal{V}'$, where observed-free cells are bordered by unobserved cells. For semantic information gain, the top-$m$ 2D grid cells with the highest semantic similarity values are used to create a Categorical distribution that is then sampled to generate data, $\mathcal{D}_s$. The frontier cells along with the samples from the Categorical distribution are then used to fit a Gaussian Mixture Model (GMM) probability distribution across the environment.

To generate candidate plans that are biased toward regions of high geometric uncertainty and semantic information gain, we use Dijkstra's algorithm on $2$D position coordinates to compute the shortest path between each observed-free cell in the occupancy grid and the robot's current state $\textbf{s}_i$, where $i$ is the MPC replanning index. From the set of all candidate plans $\mathcal{P}$, we randomly sample from the GMM for target positions. This allows the updated set of randomly sampled trajectories, $\hat{\mathcal{P}}$, to be biased toward the frontiers and the highest scoring semantic regions. This trajectory sampling procedure is shown in lines \ref{alg:frontiers} through \ref{alg:sample} of Algo.\ref{alg:vista-plan}.
While this procedure accounts for the 2D path of the robot in the environment, the viewing angles along the trajectory, $\bar{\boldsymbol{\psi}}$, are determined by pointing the robot toward the closest frontier cell or GMM mean. These viewing angles are computed to be dynamically feasible from the robot's current orientation with control limits.

After generating the candidate trajectories, we score each trajectory using (\ref{eq:vista-score}) and select the path with the highest score for the robot to track. This planning procedure is repeated in a Model-Predictive Control (MPC) loop. As the replanning loop progresses, we additionally decay $c$ in (\ref{eq:vista-score}) with parameter $\beta$ and replanning index $i$ to gradually weigh the semantic score higher than the geometric score.

\begin{algorithm}
    \caption{VISTA-Plan}\label{alg:vista-plan}
    \textbf{Input:} $\textbf{s}_{i}, \mathcal{V}, c, \beta, z$
    \begin{algorithmic}[1]
        \State $\mathcal{V}' = \texttt{FlattenVoxelGrid}(\mathcal{V})$ \label{alg:flatten}
        \State $\mathcal{D}_f = \texttt{GetFrontiers}(\mathcal{V}')$\label{alg:frontiers}
        \State $\mathcal{D}_s = \texttt{GetSemanticSamples}(\mathcal{V}')$
        \State $\mathcal{P} = \texttt{Dijkstra}(\textbf{s}_i, \mathcal{V}')$
        \State $\hat{\mathcal{P}} \sim \texttt{SampleTrajectories}(\mathcal{P}, \text{GMM}(\mathcal{D}_f, \mathcal{D}_s)) $\label{alg:sample}
        \State $\mathcal{G} = \emptyset$
        \For{$\bar{\textbf{p}} \in \hat{\mathcal{P}}$}
            \State $\bar{\boldsymbol{\psi}} = \texttt{FeasibleHeadings}(\mathcal{D}_f, \mathcal{D}_s, \bar{\textbf{p}}, \textbf{s}_i)$
            \State $\bar{\textbf{s}} \leftarrow \bar{\textbf{p}} \cup \bar{\boldsymbol{\psi}}$
            \State $\bar{\textbf{x}} \leftarrow \texttt{ConstructFullPose}(\bar{\textbf{s}}, z)$
            \State $G = \texttt{VISTAScore}(\bar{\textbf{x}}, \mathcal{V}, c)$ 
            \State $\mathcal{G} \leftarrow \mathcal{G} \cup G$
        \EndFor
        \State $ c = \beta^{i} c$
        \State $\bar{\textbf{s}}^* = \texttt{GetBestTrajectory}(\mathcal{G},\hat{\mathcal{P}})$
    \end{algorithmic}
\end{algorithm}

\section{Results}
\label{sec:results}
To demonstrate the contributions of our method, we first compare our geometric information gain method to two baseline methods: FisherRF \cite{jiang2023fisherrf} and Bayes' Rays \cite{goli2024bayes}, on static image-pose datasets that are collected from the real world. This evaluation setup allows us to directly compare our proposed information gain metric with prior work. We then incorporate semantic information and our proposed planning approach to implement the full pipeline in hardware on a quadrotor platform. On this hardware platform, we compare our method to two baselines; one using only semantic relevance and the other only geometric information gain. The semantic information-only baseline plans greedily toward the highest semantic information point in the 3DGS map, while the geometric information gain baseline is our reimplementation of RT-Guide \cite{tao2024rt}. This experiment evaluates the advantages of fusing semantic information and geometric uncertainty in robot exploration problems. Lastly, we demonstrate our full pipeline in hardware on a Boston Dynamics Spot quadruped robot to show the versatility of our method to different types of hardware platforms. We show third-person views of the Quadrotor and Spot robot in their respective testing environments in Fig.~\ref{fig:hardware}. 

\begin{figure*}[h]
    \centering
    \includegraphics[width=\textwidth, trim={1cm 13cm 0.5cm 0cm}]{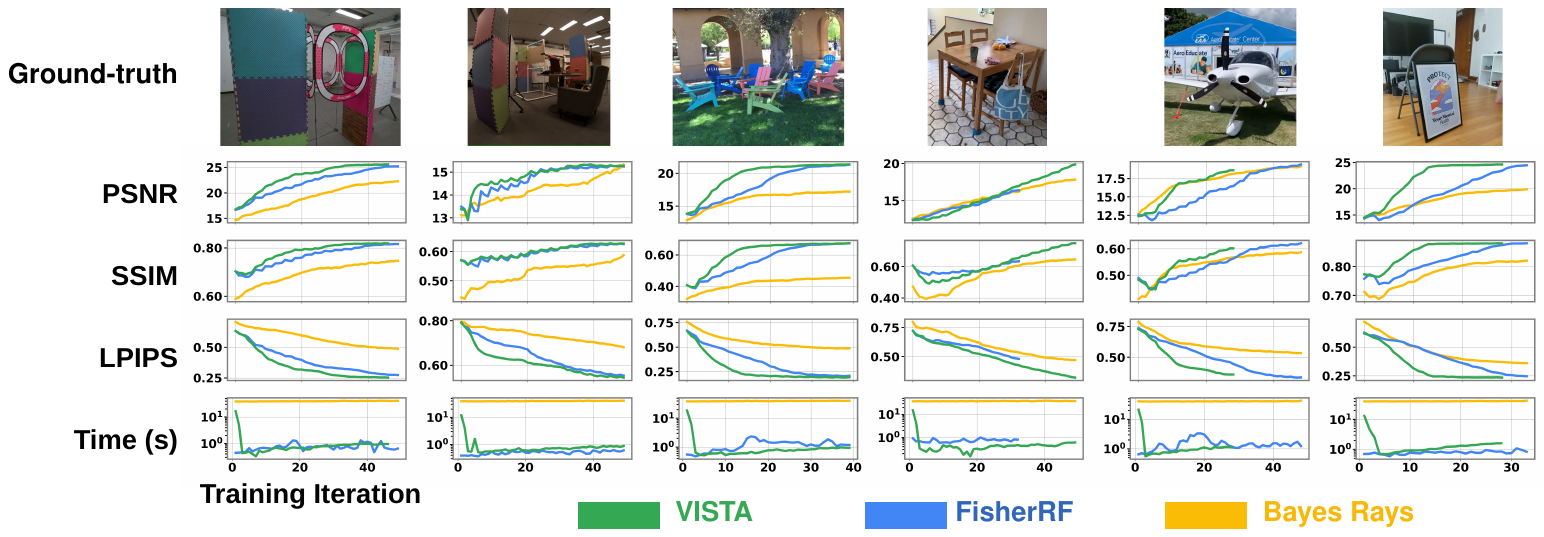}
    \caption{Our geometric information gain metric significantly outperforms baselines FisherRF and Bayes Rays in the next best view selection task for about 50K iterations in three visual reconstruction metrics and computation time. Six real scenes were used in this comparison, three from the Nerfstudio dataset $\{\textit{Plane, Kitchen, Poster}\}$ and three additional datasets $\{\textit{Flight, Clutter, Adirondacks}\}$.}
    \vspace{-1em}
    \label{fig:vista-sim-quantiative}
\end{figure*}
\subsection{Next Best View Selection Baseline Comparisons}
To evaluate our geometric information gain metric, we compare against baseline approaches FisherRF \cite{jiang2023fisherrf} and Bayes' Rays \cite{goli2024bayes}. In our baseline comparisons, we train a radiance field using a predetermined set of training views for a fixed number of iterations ($1000$). We then apply each geometric information gain metric to select a fixed number of additional views. After including these additional views, we train the models for $1000$ iterations and render images from fixed test viewpoints. We evaluate each method using the standard metrics: Peak-Signal-Noise-Ratio (PSNR), Learned Perceptuation Image Patch Similarity (LPIPS), and Structural Similarity Index Measure (SSIM). 

We evaluate each method across six scenes: three benchmark scenes in Nerfstudio (\emph{Plane}, \emph{Kitchen}, and \emph{Poster}) and three additional datasets (\emph{Flight}, \emph{Clutter}, and \emph{Adirondacks}), shown in Fig. \ref{fig:vista-sim-quantiative}.
We provide the performance results of each method in Fig. \ref{fig:vista-sim-quantiative}. We find that VISTA achieves the highest PSNR and SSIM scores and the lowest LPIPS score across all scenes. Moreover, VISTA requires the fewest number of training iterations to reach the best PSNR, SSIM, and LPIPS scores in many scenes, demonstrating the VISTA's superiority in selecting informative views compared to the other methods. For example, the best-competing method, FisherRF, requires almost twice as many training iterations to achieve the same photometric scores as VISTA, in the \emph{Poster} and \emph{Adirondacks} scenes.   
Meanwhile, VISTA requires about the same or lower computation time as the best-competing methods.
These results indicate that VISTA's information-gain metric accurately captures the information content of candidate views, compared to prior work.

\subsection{Quadrotor Hardware Experiments}

We first demonstrate our full method in hardware on a custom-built quadrotor that uses a ZED-Mini stereo camera for images, and a Jetson Orin Nano onboard computer. For pose feedback, we use an OptiTrack external motion capture system, and all 3DGS training and planning is done on a desktop computer that has an Intel(R) Core(TM) i9-13900K CPU, and an NVIDIA GeForce RTX 4090 GPU. The quadrotor receives waypoints from the offboard computer and uses an onboard state machine and PX4 flight controller for lower level control. 

We test our full system on a quadrotor platform and compare against two different baseline approaches. The first baseline solely uses the geometric information gain to quantify trajectories that are sampled uniformly throughout the environment, and is based off the work in RT-Guide \cite{tao2024rt}. Specifically, we compute the change in the means of the Gaussians in the scene, and use this signal to find the regions of the environment where the Gaussians are changing the most. Instead of using the planning method that is used in the original paper, we adapt our method to seed the GMM with these high changing Gaussians. When trajectories are sampled from the GMM and scored, they are scored by the counts of the high uncertainty and low uncertainty Gaussians visible from each view point along the sampled trajectory. In the second baseline, the method greedily plans toward the point from the 3DGS point cloud with the highest semantic similarity and points its heading along the velocity vector of the path. 

\begin{figure*}[h]
    \centering
    \includegraphics[width=\linewidth, trim={0 2cm 0 -1cm}]{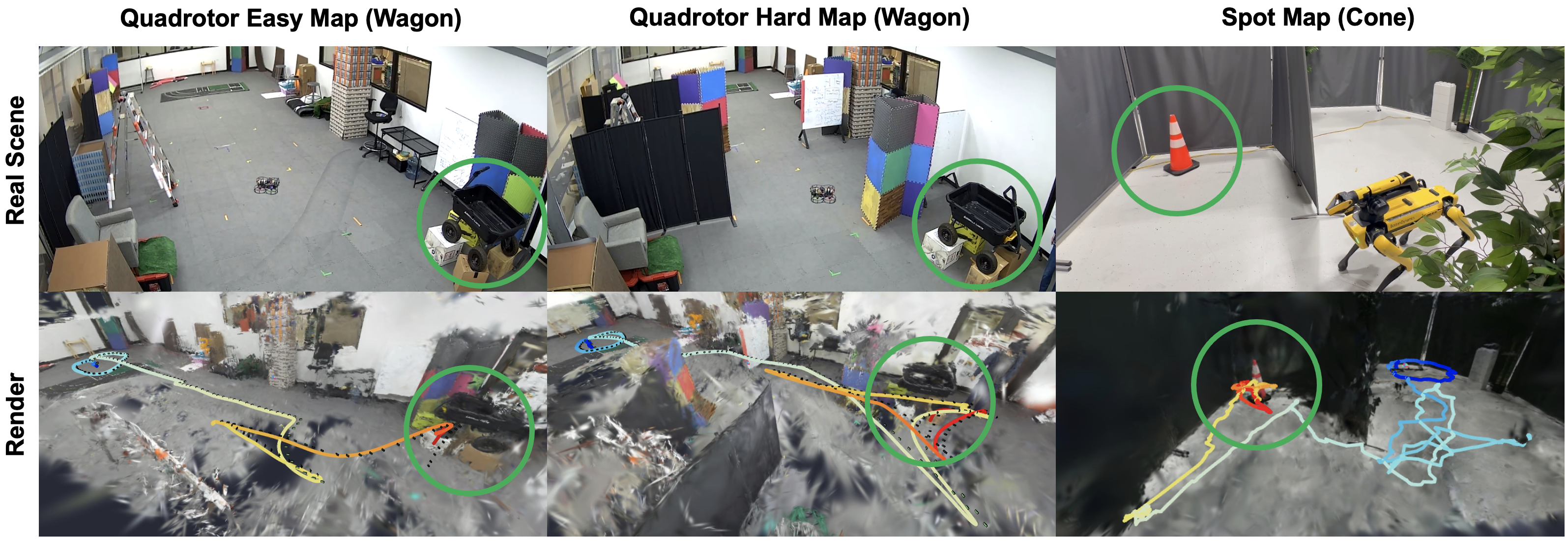}
    \caption{The top row shows our three environments and two robots, with the search object in a green circle. The second row shows an example trajectory of the robot as it executes VISTA. Trajectories are color coded from blue (beginning) to red (end). In the first two maps (columns 1 and 2), we prompt the quadrotor with the search term ``wagon."  In the last map (column 3), we prompt the spot quadruped with the term ``cone."} 
    \vspace{-1em}
    \label{fig:hardware}
\end{figure*}

We compare our full pipeline against these two baselines using three different query objects, with two different map configurations, illustrated in Fig. \ref{fig:hardware}. In the first map configuration, the query objects are not occluded in the environment. In the second map configuration, we intentionally occlude the objects from the drone. We expect most methods to be able to succeed in the first map, as the query object should be relatively easy to find and have the stopping condition trigger. In the second map, we expect methods that do not account for geometric information gain to struggle to find the query object.

We evaluate all methods on success rate (SR), time to reach (TTR), and success weighted by inverse path length (SPL), as done in \cite{gadre2023cows} and \cite{yokoyama2024vlfm}. We enforce a maximum amount of time for the quadrotor to find the query object based on battery life. Each method is tested on a query and map for two trials, totaling 12 trials for each method, six on each map. All methods are given an initialization phase, where the robot turns in a small circle about its starting pose. The results are shown in Table \ref{tab:hardware_results_easy_map}.

\begin{table*}[th]
	\centering
	\caption{Comparison metrics of time to task completion between our method and baseline methods in the Low-Occlusion Map (Easy Map Domain) and High-Occlusion Map (Hard Map Domain)}
	\label{tab:hardware_results_easy_map}
		{\begin{tabular}{c | l | c c c  | c c c  | c c c }
				\toprule
                    & & \multicolumn{3}{c |}{\emph{Ladder} (close)} & \multicolumn{3}{c |}{\emph{Sofa} (medium)} & \multicolumn{3}{c}{\emph{Wagon} (far)} \\
				Map & Methods & SR $\%$ $\uparrow$ & TTR  $\downarrow$ & SPL $\%$ $\uparrow$ & SR $\%$ $\uparrow$ & TTR  $\downarrow$ & SPL $\%$ $\uparrow$ & SR $\%$ $\uparrow$ & TTR  $\downarrow$ & SPL $\%$ $\uparrow$  \\
				\midrule
                    & RT-GuIDE \cite{tao2024rt} & \cellcolor{Goldenrod!40}50 & 154.32 & 7.19
 & \cellcolor{Green!30}100 & \cellcolor{Goldenrod!40}85.22 &  \cellcolor{Green!30}46.31 & \cellcolor{Green!30}50 & \cellcolor{Goldenrod!40}57.20 & \cellcolor{Green!30}33.52\\
                    Easy & Semantic  & \cellcolor{Green!30}100 & \cellcolor{Green!30}74.23 & \cellcolor{Green!30}31.19 
 & \cellcolor{Goldenrod!40}50 & 123.18 & 12.81  & \cellcolor{Goldenrod!40}0 & N/A & 0 \\
                    & VISTA [\textbf{ours}] & \cellcolor{Green!30}100 & \cellcolor{Goldenrod!40}83.72 & \cellcolor{Goldenrod!40}29.04 
 & \cellcolor{Green!30}100 & \cellcolor{Green!30}72.61 & \cellcolor{Goldenrod!40}38.42  & \cellcolor{Green!30}50 & \cellcolor{Green!30}56.21 & \cellcolor{Goldenrod!40}31.27 \\
 \midrule
 & RT-GuIDE \cite{tao2024rt} & \cellcolor{Goldenrod!40}50 & \cellcolor{Goldenrod!40}145.25 &  \cellcolor{Goldenrod!40}10.59
 & \cellcolor{Goldenrod!40}0 & \cellcolor{Goldenrod!40}N/A & \cellcolor{Goldenrod!40}0   & \cellcolor{Goldenrod!40}0 & \cellcolor{Goldenrod!40}N/A & \cellcolor{Goldenrod!40}0\\
                    Hard & Semantic  & \cellcolor{Goldenrod!40}50 & 159.21 & 8.97  
 & \cellcolor{Goldenrod!40}0 & \cellcolor{Goldenrod!40}N/A & \cellcolor{Goldenrod!40}0   & \cellcolor{Goldenrod!40}0 & \cellcolor{Goldenrod!40}N/A & \cellcolor{Goldenrod!40}0 \\
                    & VISTA [\textbf{ours}] & \cellcolor{Green!30}100 & \cellcolor{Green!30}141.69 & \cellcolor{Green!30}16.26  
 & \cellcolor{Green!30}100 & \cellcolor{Green!30}123.24 & \cellcolor{Green!30}38.92   & \cellcolor{Green!30}100 & \cellcolor{Green!30}109.89 & \cellcolor{Green!30}33.82 \\
				\bottomrule
		\end{tabular}}
    \vspace{-1em}
\end{table*}

Through these experiments, we find that all methods have some successes on the easy low-occlusion map domain. Our method has the highest success rate on this map with an $83.33\%$ success rate over the RT-Guide baseline success rate of $66.67\%$, and semantic baseline success rate of $50\%$. On the more challenging map domain, we find that our method has a significant improvement over the baseline methods, where our method has a $100\%$ success rate while both baselines each have a $16.67\%$ success rate. The results suggest that our method is able to outperform both baselines on both maps because we reason about both semantic and geometric information gain.
\subsection{Spot Quadruped Hardware Experiments}
For our second hardware platform, we use a Boston Dynamics Spot quadruped robot fitted with RGB-D cameras and onboard odometry. In these experiments, only the front two cameras on the Spot robot body are used to train the 3DGS map. The offboard computer is equipped with a 4.2 GHz AMD Ryzen 7 7800X3D CPU and an NVIDIA GeForce RTX 4090 (24GB memory). We communicate with the Boston Dynamics Spot via the SDK and execute the task-aware plans using desired waypoint control. Qualitative results of our method on the Spot robot are shown in Fig. \ref{fig:hardware}.

\section{Conclusion}
\label{sec:conclusion}
In this work, we present an information gain metric combining both geometric information as well as semantic gain, and demonstrate how these metrics can be used on a real-time hardware platform to simultaneously map an environment and find an object in the map specified through natural language. We find that our method, VISTA, is fast enough to run real-time on robot hardware supported by an offboard GPU workstation, and show that by using a combined metric for semantic and geometric information gain, we can more quickly focus on areas of the map that have higher relevance to the search task. We compared our geometric information gain metric to previously published baseline methods using pre-collected datasets of images and poses, and demonstrated VISTA on two different hardware platforms in exploration tasks with varying map difficulty.
\paragraph*{Limitations} VISTA uses CLIP features, which are known to mostly encode object-centric semantics.  This limits VISTA to performing object search tasks.  VISTA cannot disambiguate between multiple instances of the same object. 
More grammatically sophisticated VLM embeddings could enable more nuanced search tasks (avoiding danger, using landmark hints, more dynamic tasks like target following).  VISTA currently requires offboard GPU compute, limiting its range and potential for field operations. VISTA also requires that the robot has its own low-level control and localization stack.  In our experiments, this is accomplished with a motion capture system for the quadrotor, and onboard SLAM system for the spot quadruped.    

\bibliographystyle{IEEEtran}
\bibliography{references}

\end{document}